\def\year{2021}\relax
\documentclass[letterpaper]{article} 
\usepackage{aaai21}  
\usepackage{times}  
\usepackage{helvet} 
\usepackage{courier}  
\usepackage[hyphens]{url}  
\usepackage{graphicx} 
\usepackage{graphicx}  
\usepackage{natbib}  
\usepackage{caption} 

\title{Weakly Supervised Temporal Action Localization Through\\Learning Explicit Subspaces for Action and Context}
 \author{
    Ziyi Liu\textsuperscript{\rm 1},
    Le Wang\textsuperscript{\rm 1}\thanks{Corresponding author.},
    Wei Tang\textsuperscript{\rm 2},
    Junsong Yuan\textsuperscript{\rm 3},
    Nanning Zheng\textsuperscript{\rm 1},
    Gang Hua\textsuperscript{\rm 4}
 }
 \affiliations{

    \textsuperscript{\rm 1}Institute of Artificial Intelligence and Robotics, Xi`an Jiaotong University~\\
    \textsuperscript{\rm 2}University of Illinois at Chicago~
    \textsuperscript{\rm 3}The State University of New York at Buffalo~
    \textsuperscript{\rm 4}Wormpex AI Research \\
    liuziyi@stu.xjtu.edu.cn,
    \{lewang, nnzheng\}@mail.xjtu.edu.cn,\\
    tangw@uic.edu,
    jsyuan@buffalo.edu,
    ganghua@gmail.com
 }

\newcommand{\TAL}{TAL}
\newcommand{\lzynet}{CleanNet}

\newcommand{\wtal}{WS-TAL}

\def\ie{\emph{i.e}.}

\def\eg{\emph{e.g}.}
\usepackage{times}
\usepackage{epsfig}
\usepackage{graphicx}
\usepackage{amsmath}
\usepackage{amssymb}
\usepackage{subcaption}
\usepackage{placeins}
\usepackage{multirow}
\usepackage{array}
\usepackage{diagbox}
\usepackage{multirow}
\usepackage[switch]{lineno}  %
\begin{document}
\maketitle

\begin{abstract}
Weakly-supervised Temporal Action Localization (WS-TAL) methods learn to localize temporal starts and ends of action instances in a video under only video-level supervision. Existing WS-TAL methods rely on deep features learned for action recognition. However, due to the mismatch between classification and localization, these features cannot distinguish the frequently co-occurring contextual background, \ie, the context, and the actual action instances. We term this challenge action-context confusion, and it will adversely affect the action localization accuracy. To address this challenge, we introduce a framework that learns two feature subspaces respectively for actions and their context. By explicitly accounting for action visual elements, the action instances can be localized more precisely without the distraction from the context. To facilitate the learning of these two feature subspaces with only video-level categorical labels, we leverage the predictions from both spatial and temporal streams for snippets grouping. In addition, an unsupervised learning task is introduced to make the proposed module focus on mining temporal information. The proposed approach outperforms state-of-the-art WS-TAL methods on three benchmarks, \ie, THUMOS14, ActivityNet v1.2 and v1.3 datasets.
\end{abstract}

\section{Introduction}
\begin{figure*}[t]
  \begin{center}
  \includegraphics[width=0.96\textwidth]{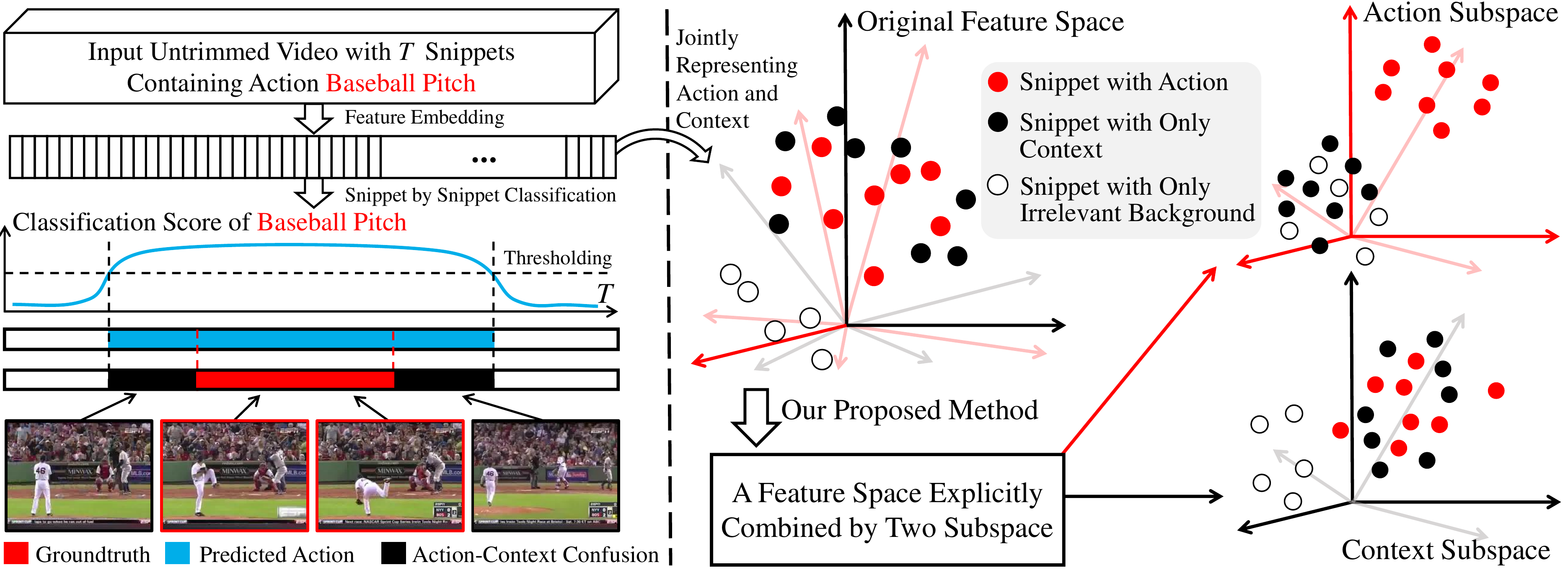}\\
  \caption{Illustration of a common WS-TAL pipeline with action-context confusion (left) and the main idea of our method (right). The original feature space jointly represents action and context visual elements, resulting in confusion of actions (red dots) and context (black dots). The proposed method learns an action subspace and a context subspace.
  By representing action visual elements in a separate feature subspace, we can alleviate the distraction from the context during temporal localization.
  In the action subspace, we require the black dots to be similar to snippets with only irrelevant background (circles) and distinct from red dots. While in the context subspace, the black dots are similar to red dots due to the spatial co-occurrence between action and context. We require both the black and red dots are distinct from circles in the context subspace.
  }\label{fig:fig1}
  \end{center}
\end{figure*}

\noindent Temporal action localization (\TAL) means to localize temporal starts and ends of action instances in an untrimmed video. It is a fundamental task in video understanding and has numerous applications such as intelligent surveillance, video retrieval and video summarization~\cite{Dan2013Action,surveyDeepAction,Kang2016Review}.
Since untrimmed videos can contain irrelevant content and multiple action instances, training fully supervised TAL methods~\cite{chao2018rethinking,lin2018bsn,lin2019bmn,MGG,GTAD} requires temporal boundary annotations of action instances. However, obtaining such annotations is time-consuming, which drives the research of weakly-supervised temporal action localization (WS-TAL) methods.

WS-TAL learns to localize action instances with only video-level categorical labels provided during training. However, the mismatch between this video-level classification supervision and the temporal localization task makes it difficult to distinguish the actual action instances from their frequently co-occurring contextual background (termed \emph{context}). For example, the action baseball pitch commonly co-occurs with the context baseball field both temporally and spatially. We term this challenge \emph{action-context confusion}.

Most existing methods train a video-level classifier and apply it snippet by snippet to achieve temporal action localization, as shown in Figure~\ref{fig:fig1}. Unfortunately, this pipeline can only exclude the irrelevant background but not the relevant context from the localization results.
While the former is class-agnostic, the latter is action-specific: the context of an action seldom co-occurs with other actions.
Therefore, irrelevant background is useless to classification, but context can provide important classification cues. For example, recognizing the baseball field will prevent the video from being misclassified as diving.
Therefore, features trained on video-level categorical labels will look for contextual cues, and the entire learned feature space will represent action and context jointly. Consequently, snippets with only context are frequently mistaken as part of the action instance.

The goal of this paper is to address the challenge of action-context confusion in WS-TAL. Our idea is to learn a feature space which can be explicitly split into two (feature) subspaces, \ie, the action subspace and the context subspace, as shown in Figure~\ref{fig:fig1}.
After representing action visual elements in a feature subspace separated from context visual elements, the target action instances can be localized considering only action elements represented in the action subspace. Therefore, the distraction from the context is alleviated, in spite of its temporal and spatial co-occurrence with action.

However, learning explicit action and context subspaces with insufficient supervision is nontrivial.
The key challenge is how to collect snippets with action/only context/only irrelevant background (\ie, red dots/black dots/circles in Figure~\ref{fig:fig1}). We take advantage of the two-stream video representation, \ie, spatial (RGB) and temporal (optical flow) streams for snippets grouping.
Specifically, we consider snippets receiving consistent positive/negative predictions from both streams as snippets with action/only irrelevant background; snippets receiving inconsistent predictions from two streams are regarded as snippets with only context. Intuitively, snippets containing typical static scenes (such as a baseball field) provide useful appearance features but lack meaningful motion. Thus, they are considered positive by the spatial stream but negative by the temporal stream. Similarly,
snippets containing typical non-action motion (such as the splash after diving or camera movement) are considered positive by the temporal stream but negative by the spatial stream.
Besides, to further remedy the lack of supervision, we introduce an additional unsupervised training task. Together with a new temporal residual module, it makes our method focus on mining useful temporal information.

In summary, our contributions are as follows.
(1) We propose to address the challenge of action-context confusion for WS-TAL by learning two feature subspaces, \ie, the action subspace and the context subspace. It helps exclude the contextual background from the localization results.
(2) We introduce a temporal residual module and an unsupervised training task to make our method focus on mining useful temporal information.
(3) Our method achieves state-of-the-art performance on three benchmarks.

\section{Related Work}
\subsubsection{TAL with Full Supervision}

TAL with full supervision requires temporal boundary annotations of the target action class during training. Similar to 2D object detection, the TAL task can be formulated as 1D object detection. Following its success in 2D object detection~\cite{girshick2015fast,ren2017faster}, the two-stage pipeline is leveraged by fully-supervised \TAL~methods~\cite{escorcia2016daps,shou2016temporal,buch2017sst,gao2017turn,xu2017RC3D,zhao2017temporal,chao2018rethinking,lin2018bsn}.
Recently, instead of viewing temporal action proposals individually, the dependencies among them are considered. BMN~\cite{lin2019bmn} introduces the boundary-matching mechanism to capture the temporal context of each proposal. Graph Convolutional Networks (GCN) is a popular tool to capture the proposal-proposal relations, as proposed in P-GCN~\cite{zeng2019graph}, G-TAD~\cite{xu2020g} and AGCN~\cite{li2020graph}.

\subsubsection{TAL with Weak Supervision}\label{sec:TALweakSuper}
WS-TAL methods focus on achieving TAL with only video-level categorical labels. Without temporal boundary annotations for explicit supervision, the attention mechanism is widely used for distinguishing action and action-agnostic background.
UntrimmedNet~\cite{wang2017untrimmednets} formulates the attention mechanism as a soft selection module to localize target action.
STPN~\cite{nguyen2017weakly} proposes a sparsity loss to improve UntrimmedNet and leverage multi-layer structure for attention learning.
W-TALC~\cite{WTALC} proposes a co-activity loss to enforce the feature similarity among localized instances.
AutoLoc~\cite{shou2018autoloc} designs an ``outer-inner-contrastive loss'' for proposal evaluation, to facilitate the temporal boundary regression.
CleanNet~\cite{lzyiccv} designs a ``contrast score'' by leveraging temporal contrast in SCPs to achieve end-to-end
training of localization.
BM~\cite{nguyen2019weakly} achieves better TAL performance via background modelling and other unsupervised losses to guide the attention.
CMSC~\cite{Liu_2019_CVPR} exploits the action-context confusion challenge by regarding frames with low optical flow intensity as background.

However, the existing WS-TAL methods did not realize the learned feature represents action and context visual elements jointly,
resulting the action-context confusion when temporally localizing the target action instances. By learning explicit subspaces for action and context, the proposed method can better avoid the distraction from the context, in spite of its temporal and spatial co-occurrence with action.
\begin{figure*}[t]
  \begin{center}
  \includegraphics[width=1\textwidth]{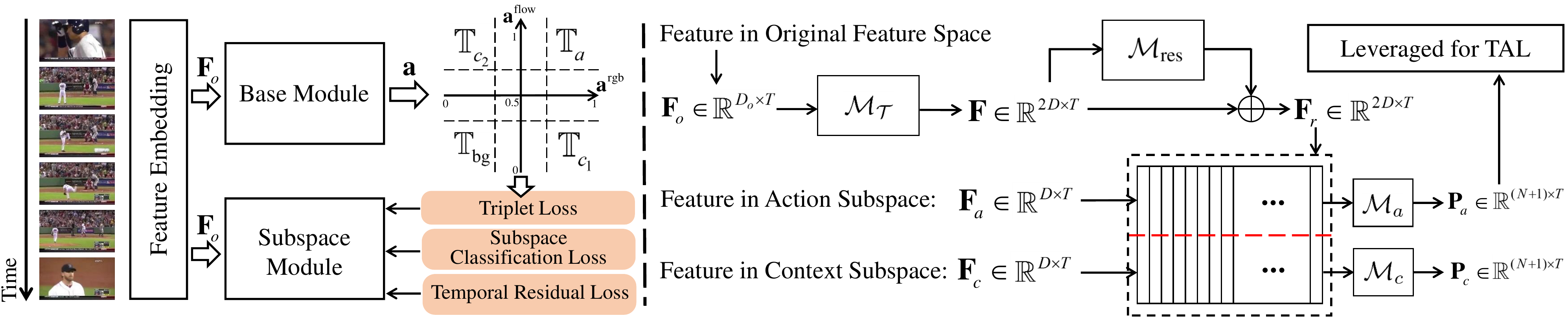}\\
  \caption{An overview of our method (left) and the detailed architecture of the proposed Subspace module (right). Left: our proposed method is mainly composed of three modules, \ie, feature embedding, Base module and Subspace module. Leveraging the Base module, we divide snippets into four sets, \ie, $\mathbb{T}_a,\mathbb{T}_{c_1},\mathbb{T}_{c_2}$ and $\mathbb{T}_\textrm{bg}$ in Eq. (\ref{eq11}-\ref{eq14}). Right: Using the Subspace module, the snippet-level features in action and context subspaces are obtained. Only the output of action subspace is used for localization to alleviate the distraction from context.
  }\label{fig:framework}
  \end{center}
\end{figure*}
\section{Proposed Method}
The overview of the proposed architecture is shown in the left part of Figure~\ref{fig:framework}.
Identical architectures are applied for both spatial (RGB) and temporal (optical flow) streams. For notation simplicity, we omit the stream superscript without causing confusion in Section 3.1-3.4.

\subsection{Feature Embedding}\label{sec:feature}
Given an untrimmed video, we first divide it into $T$ non-overlapping snippets, which are the inputs of the feature embedding module. The outputs are the corresponding features of $T$ snippets, denoted as $\mathbf{F}_o \in \mathbb{R}^{D_o \times T}$. $\mathbf{F}_o(t)\in \mathbb{R}^{D_o}$ is the feature of the $t$-th snippet $s_t$. We term the feature space of $\mathbf{F}_o$ ``the original feature space'', as shown in Figure~\ref{fig:fig1}.

It is worth noting that the original feature space of snippet-level features $\mathbf{F}_o$ jointly represents the visual elements of action and context, because they frequently co-occur both temporally and spatially and only video-level category labels are available during training. The Subspace module will address this issue as detailed in Section~\ref{sec:sub}.

\subsection{Base Module}
The Base module performs video-level classification with the attention mechanism. Similar pipelines have been explored in~\cite{nguyen2017weakly, nguyen2019weakly}. This part is not included in our contribution.

After obtaining $\mathbf{F}_o$, the attention mechanism leverages $\mathbf{F}_o$ to obtain snippet-level predictions (SAPs):
\begin{equation}\label{eq:sap_s}
\mathbf{a}(t) = \mathcal{M}_{\textrm{att}}(\mathbf{F}_o(t),\phi_{\textrm{att}}),~~\mathbf{a}\in \mathbb{R}^{1 \times T},\\
\end{equation}
where $\mathcal{M}_{\textrm{att}}$ is a combination of a fully connected (FC) layer and a sigmoid function. $\phi_{\textrm{att}}$ is the corresponding learnable parameters. Afterwards, the video-level foreground and background features are fused by temporal weighted average pooling following~\cite{nguyen2019weakly}:
\begin{equation}\label{fg}
\mathbf{f}_\textrm{fg} =\frac{1}{T}\sum_{t=1}^{T}\mathbf{a}(t)\mathbf{F}_o(t),~\mathbf{f}_\textrm{fg} \in \mathbb{R}^{D_o},
\end{equation}
\begin{equation}\label{bg}
\mathbf{f}_\textrm{bg} =\frac{1}{T}\sum_{t=1}^{T}(1-\mathbf{a}(t))\mathbf{F}_o(t),~\mathbf{f}_\textrm{bg} \in \mathbb{R}^{D_o}.
\end{equation}

After obtaining video-level features, video-level predictions are obtained as
\begin{linenomath}\begin{align}\label{eq:p_fgbg}
&\mathbf{p}_\textrm{fg}=\mathcal{M}_\textrm{cls}(\mathbf{f}_\textrm{fg},\phi_{\textrm{cls}}),~\mathbf{p}_\textrm{fg} \in \mathbb{R}^{(N+1)},\\
&\mathbf{p}_\textrm{bg}=\mathcal{M}_\textrm{cls}(\mathbf{f}_\textrm{bg},\phi_{\textrm{cls}}),~\mathbf{p}_\textrm{bg} \in \mathbb{R}^{(N+1)},
\end{align}\end{linenomath}
where $\mathcal{M}_\textrm{cls}$ is implemented as an FC layer to get classification results. $(N+1)$ indicates the total number of action classes including the background class.

Applying $\mathcal{M}_\textrm{cls}$ to features of each snippet $\mathbf{F}_o(t)$, the snippet-level classification predictions are obtained as
\begin{equation}
\mathbf{P}_o(t) = \mathcal{M}_\textrm{cls}(\mathbf{F}_o(t),\phi_{\textrm{cls}}), ~~\mathbf{P}_o \in \mathbb{R}^{(N+1) \times T}.
\end{equation}

\subsection{Subspace Module}\label{sec:sub}
The goal of the Subspace module is to transform the original feature space to a feature space explicitly
combined by two subspaces, \ie, the action subspace and the context subspace. The architecture is shown in the right part of Figure~\ref{fig:framework}.
First, we transform the feature of the $t$-th snippet $\mathbf{F}_o(t)$ as
\begin{equation}\label{eq:trans}
\mathbf{F}(t)=\mathcal{M}_{\mathcal{T}}(\mathbf{F}_o(t), \phi_{\mathcal{T}}),~\mathbf{F} \in \mathbb{R}^{2D \times T},
\end{equation}
where $\mathcal{M}_{\mathcal{T}}$ is implemented as an FC layer with a ReLU activation and $\phi_{\mathcal{T}}$ denotes its trainable parameters. The combined feature space is $2D$ dimensional, with $D$ dimensions for each subspace.
Such a transformation is snippet independent. To consider snippet contextual information, a temporal residual module (T-ResM) is leveraged to enhance the learned features:
\begin{equation}\label{eq:t_res}
\mathbf{F}_r=\mathbf{F} + \mathcal{M}_\textrm{res}(\mathbf{F},\phi_\textrm{res}),~\mathbf{F}_r \in \mathbb{R}^{2D \times T},
\end{equation}
where $\mathcal{M}_\textrm{res}$ is implemented using two stacked temporal 1d-convolutional layers, $\phi_\textrm{res}$ represents the learnable parameters.
Note that the T-ResM is optional. With only video-level categorical labels available during training, additional layers such as $\mathcal{M}_\textrm{res}$ will adversely affect the training process due to the lack of supervision. Therefore, an unsupervised training task is introduced to facilitate the training of $\mathcal{M}_\textrm{res}$, which will be detailed in Section~\ref{sec:train}.

After obtaining features in the combined feature space (\ie, $\mathbf{F}_r$), features in the action subspace $\mathbf{F}_a \in \mathbb{R}^{D \times T}$ and features in the context subspace $\mathbf{F}_c \in \mathbb{R}^{D \times T}$ are obtained by directly splitting $2D$ channels into two parts, each with $D$ channels, \ie,
\begin{equation}\label{}
\mathbf{F}_r=[\mathbf{F}_a, \mathbf{F}_c],
\end{equation}
where $[\cdot]$ indicates vector concatenation.

Subsequently, for the $t$-th snippet, we can obtain its predictions from the action subspace ($\mathbf{P}_a(t)\in \mathbb{R}^{(N+1)}$) and the context subspace ($\mathbf{P}_c(t)\in \mathbb{R}^{(N+1)}$) as
\begin{linenomath}\begin{align}\label{}
\mathbf{P}_a(t)=\mathcal{M}_a(\mathbf{F}_a(t), \phi_a),~
\mathbf{P}_c(t)=\mathcal{M}_c(\mathbf{F}_c(t), \phi_c),
\end{align}\end{linenomath}
where $\mathcal{M}_a$ and $\mathcal{M}_c$ represent the classification layers for action and context subspaces with trainable parameters $\phi_a$ and $\phi_c$, respectively.

\noindent \textbf{Summary of the proposed architecture.}
For Base module, in the training stage, its outputs are (1) $\mathbf{p}_\textrm{fg} \in \mathbb{R}^{(N+1)}$ and $\mathbf{p}_\textrm{bg} \in \mathbb{R}^{(N+1)}$ for Base module training; (2) $\mathbf{a} \in \mathbb{R}^{1 \times T}$ for Subspace module training. In the testing stage, the outputs of Base module are (1) $\mathbf{a}$ for temporal action proposal (TAP) generation; (2) $\mathbf{P}_o \in \mathbb{R}^{(N+1) \times T}$ for TAP evaluation.

For Subspace module, in the training stage, its outputs are (1) $\mathbf{P}_a \in \mathbb{R}^{(N+1) \times T}$ and $\mathbf{P}_c \in \mathbb{R}^{(N+1) \times T}$ for the training of Subspace module; (2) $\mathbf{F}_r \in \mathbb{R}^{2D \times T}$ for the training of T-ResM. In the testing stage, the output is $\mathbf{P}_a$ for both TAP generation and evaluation. The context subspace is only used for facilitating the training process and not considered for performing TAL.

The above architecture is applied for both spatial and temporal streams. Adding the superscript ``\textrm{rgb}'' or ``\textrm{flow}'' to notations above, we can obtain the notations for the spatial or temporal stream. This also applies the training process below.

\subsection{Training Process for Classification}\label{sec:train}
This section introduces how to learn explicit action and context subspaces with only video-level labels.
The training of the Base module has been well explored and is not our main contribution.

\subsubsection{An Overview of the Training of Subspace Module}
Our training objective is to ensure the obtained action/context subspace and the feature $\mathbf{F}_r \in \mathbb{R}^{2D \times T}$ have three properties.
The \textbf{first} property is about the similarities among features in different subspaces, guided by the triplet loss $L_t$.
The \textbf{second} property is about the classification predictions of features in different subspaces, guided by the subspace classification loss $L_s$.
The \textbf{third} property is that $\mathbf{F}_r \in \mathbb{R}^{2D \times T}$ should contain temporal information, guided by an optional temporal residual loss $L_r$. The three properties and their corresponding losses are detailed below. 

\subsubsection{Preparation before Loss Calculation}
The sets of snippets with action/only context/only irrelevant background (denoted as $\mathbb{S}_a$/$\mathbb{S}_c$/$\mathbb{S}_\textrm{bg}$) are required for calculation of losses. Specifically, we first collect four snippet index sets as
\begin{linenomath}\begin{align}\label{eq11}
&\mathbb{T}_{a~}=\{t~|~\mathbf{a}^\textrm{rgb}(t)>\theta_h ~\&~ \mathbf{a}^\textrm{flow}(t)>\theta_h \},\\
&\mathbb{T}_{c_1}=\{t~|~\mathbf{a}^\textrm{rgb}(t)>\theta_h ~\&~ \mathbf{a}^\textrm{flow}(t)<\theta_l\},\\
&\mathbb{T}_{c_2}=\{t~|~\mathbf{a}^\textrm{rgb}(t)<\theta_l ~\&~ \mathbf{a}^\textrm{flow}(t)>\theta_h\},\\\label{eq14}
&\mathbb{T}_\textrm{bg}=\{t~|~\mathbf{a}^\textrm{rgb}(t)<\theta_l ~\&~ \mathbf{a}^\textrm{flow}(t)<\theta_l\},
\end{align}\end{linenomath}
where $\theta_h$ and $\theta_l$ are the high and low thresholds, defined as
\begin{linenomath}\begin{align}\label{eq:gap}
\theta_h = 0.5+ \alpha,~\theta_l = 0.5- \alpha,
\end{align}\end{linenomath}
where $\alpha$ is the hyper-parameter controlling the gap between $\theta_h$ and $\theta_l$.
Subsequently, $\mathbb{S}_a$ (or $\mathbb{S}_\textrm{bg}$) is obtained by collecting snippets with consistent positive (or negative) predictions from both streams, \ie, $\mathbb{S}_a=\{s_t|t \in \mathbb{T}_{a}\}$ (or $\mathbb{S}_\textrm{bg}=\{s_t|t \in \mathbb{T}_\textrm{bg}\}$). Similarly, $\mathbb{S}_c$ is obtained by collecting snippets with inconsistent predictions from two streams as $\mathbb{S}_c=\{s_t|t \in \mathbb{T}_{c_1} \cup \mathbb{T}_{c_2}\}$.

\subsubsection{First Property and Triplet Loss}
The first property: In the action subspace, the representation of $s_t \in \mathbb{S}_c$ should be similar with that of $s_t \in \mathbb{S}_\textrm{bg}$, because both sets of snippets contain no action. Meanwhile, the representation of $s_t \in \mathbb{S}_c$ should be distinct from that of $s_t \in \mathbb{S}_a$.
On the contrary, in the context subspace, the representation of $s_t \in \mathbb{S}_c$ should be distinct from that of $s_t \in \mathbb{S}_\textrm{bg}$. Meanwhile, the representation of $s_t \in \mathbb{S}_c$ should be similar to that of $s_t \in \mathbb{S}_a$ because of the spatial co-occurrence between action and context. This property can be naturally guided by the triplet loss.

In the action subspace, we obtain three video-level features representing action, context and irrelevant background snippets as
\begin{linenomath}\begin{align}\label{}
&\mathcal{A}_a=\frac{1}{|{\mathbb{T}_a}|}\sum_{t \in {\mathbb{T}_a}}\mathbf{F}_a(t),\\
&\mathcal{A}_c=\frac{1}{|{\mathbb{T}_c}|}\sum_{t \in {\mathbb{T}_c}}\mathbf{F}_a(t),\\
&\mathcal{A}_\textrm{bg}=\frac{1}{|{\mathbb{T}_\textrm{bg}}|}\sum_{t \in {\mathbb{T}_\textrm{bg}}}\mathbf{F}_a(t),
\end{align}\end{linenomath}
where $|\cdot|$ denotes the number of elements. Similarly,
in the context subspace, we obtain video-level features representing action, context and irrelevant background snippets as
\begin{linenomath}\begin{align}\label{}
&\mathcal{C}_a=\frac{1}{|{\mathbb{T}_a}|}\sum_{t \in {\mathbb{T}_a}}\mathbf{F}_c(t),\\
&\mathcal{C}_c=\frac{1}{|{\mathbb{T}_c}|}\sum_{t \in {\mathbb{T}_c}}\mathbf{F}_c(t),\\
&\mathcal{C}_\textrm{bg}=\frac{1}{|{\mathbb{T}_\textrm{bg}}|}\sum_{t \in {\mathbb{T}_\textrm{bg}}}\mathbf{F}_c(t).
\end{align}\end{linenomath}
Afterwards, we leverage the triplet loss as
\begin{linenomath}\begin{align}\label{}
L_{t} = \textrm{max}(\bar{d}(\mathcal{A}_c,\mathcal{A}_\textrm{bg}),\bar{d}(\mathcal{A}_c,\mathcal{A}_a)+m,0) + \nonumber\\
~~~\textrm{max}(\bar{d}(\mathcal{C}_c,\mathcal{C}_a)-\bar{d}(\mathcal{C}_c,\mathcal{C}_\textrm{bg})+m,0),~~
\end{align}\end{linenomath}
where $\bar{d}(\mathbf{p},\mathbf{q})$ is the Euclidean distance between the $\ell_2$ normalized $\mathbf{p}$ and $\mathbf{q}$. $m$ is the margin set as $1$.

\subsubsection{Second Property and Subspace Classification Loss}
The second property: In the action subspace, $\forall s_t \in \mathbb{S}_a$ should be predicted as having target action class (Eq.~(\ref{eq25})). While $\forall s_t \in \mathbb{S}_c \cup \mathbb{S}_\textrm{bg}$ should be predicted as the background class (Eq.~(\ref{eq26})). In the context subspace, $\forall s_t \in \mathbb{S}_c $ should be classified as the target action class (Eq.~(\ref{eq27})), while $\forall s_t \in \mathbb{S}_\textrm{bg}$ should be predicted as the background class (Eq.~(\ref{eq28})). No explicit constraint is imposed on the predictions of $s_t \in \mathbb{S}_a $ in the context subspace.

To obtain the subspace classification loss $L_s$, the snippet-level classification labels of predictions from action and context subspaces (\ie, $\mathbf{P}_a \in \mathbb{R}^{(N+1) \times T}$ and $\mathbf{P}_c \in \mathbb{R}^{(N+1) \times T}$) are required.
Following the second property, we assign the labels to $\mathbf{P}_a$ and $\mathbf{P}_c$ as
\begin{linenomath}\begin{align}\label{eq25}
\forall t \in \mathbb{T}_a,&~~\mathbf{P}_a(t)|_n \gets 1,~~\mathbf{P}_a(t)|_0 \gets 0,\\\label{eq26}
\forall t \in \mathbb{T}_c \cup \mathbb{T}_\textrm{bg},&~~\mathbf{P}_a(t)|_n\gets0,~~\mathbf{P}_a(t)|_{0}\gets1,\\\label{eq27}
\forall t \in \mathbb{T}_c,&~~\mathbf{P}_c(t)|_n\gets1,~~\mathbf{P}_c(t)|_0\gets0,\\\label{eq28}
\forall t \in \mathbb{T}_\textrm{bg},&~~\mathbf{P}_c(t)|_{n}\gets0,~~\mathbf{P}_c(t)|_{0}\gets1,
\end{align}\end{linenomath}
where the video is annotated as having the $n$-th action class and $|_n$ indicates the classification prediction of the $n$-th action. The $0$-th class indicates the background class.

After assigning the predictions with binary expectations as Eq. (\ref{eq25}-\ref{eq28}), we can train the Subspace module using a binary logistic regression loss $L_s$. 

\subsubsection{Third Property and Temporal Residual Loss}
The third property: $\mathbf{F}_r$ should contain temporal information as T-ResM is the only module that can provide cross-snippet information in the Subspace module.
However, we observe that without additional constraints, temporal 1d-convolutional layers cannot enhance the temporal features. So we introduce an auxiliary training task to ensure that property, and the temporal residual loss $L_r$ is the corresponding loss of this task.

Specifically, the auxiliary unsupervised task is ``snippet-level four-class classification''. Regarding snippet indexes belonging to different index sets as different classes, we collect four classes indicated by Eq. (\ref{eq11}-\ref{eq14}). Afterwards, we perform the four-class prediction as
\begin{equation}
\mathbf{P}_r(t) = \mathcal{M}_r(\mathbf{F}_r(t),\phi_{r}), ~~\mathbf{P}_r \in \mathbb{R}^{4 \times T},
\end{equation}
where $\mathcal{M}_r$ is a classification layer only used for this additional task. $L_r$ is the corresponding cross-entropy loss:
\begin{equation}\label{}
L_r=-\frac{1}{|\mathbb{T'}|}\sum_{t\in \mathbb{T'}}\textrm{log}(\mathbf{P}_r(t)|_n),
\end{equation}
where $\mathbb{T}'=\{\mathbb{T}_a\cup\mathbb{T}_{c_1}\cup\mathbb{T}_{c_2}\cup\mathbb{T}_\textrm{bg}\}$, and $n=0/1/2/3$ if $t\in \mathbb{T}_a/\mathbb{T}_{c_1}/\mathbb{T}_{c_2}/\mathbb{T}_\textrm{bg}$. This prediction involves cross stream information. To achieve this goal, $\mathbf{F}_r$ is required to account for snippet contextual information. Since the Subspace module of each stream is trained independently and T-ResM is the only module that can provide cross-snippet information, minimizing $L_r$ can guide the T-ResM to focus on mining snippet contextual information.

Finally, the total loss for the training of the Subspace module is
$L=L_{t}+L_{s}+L_{r}$.

\subsection{Testing Process for TAL}
The TAL results can be achieved using the outputs from the Base module or the Subspace module in the testing process. Firstly the outputs from two streams are fused by weighted average with a hyper-parameter $\beta$, \ie,
\begin{linenomath}\begin{align}\label{eq:beta0}
&\bar{\mathbf{a}}=\beta\mathbf{a}^\textrm{rgb}+(1-\beta)\mathbf{a}^\textrm{flow},~\bar{\mathbf{a}} \in \mathbb{R}^{1\times T},\\
&\bar{\mathbf{P}}_o=\beta\mathbf{P}_o^\textrm{rgb}+(1-\beta)\mathbf{P}_o^\textrm{flow},~\bar{\mathbf{P}}_o \in \mathbb{R}^{(N+1)\times T},\\\label{eq:beta1}
&\bar{\mathbf{P}}_a=\beta\mathbf{P}^\textrm{rgb}_a+(1-\beta)\mathbf{P}^\textrm{flow}_a,~\bar{\mathbf{P}}_a \in \mathbb{R}^{(N+1)\times T}.
\end{align}\end{linenomath}

To achieve TAL, there are two necessary steps, \ie, temporal action proposal (TAP) generation and TAP evaluation.
With only Base module, the TAP generation is achieved by thresholding $\bar{\mathbf{a}}$ using $0.5$ ($C_1$ in Table~\ref{table:parts}). Given a TAP, its confidence score containing the $n$-th action is evaluated by Outer-Inner-Contrastive~\cite{shou2018autoloc} as
\begin{linenomath}\begin{align}\label{eq:oic}
&s(t_s,t_e, n, \bar{\mathbf{P}}_o)=\textrm{mean}(\bar{\mathbf{P}}_o(t_s:t_e)|_n)-\nonumber \\
&~~~~~~~~~~\textrm{mean}([\bar{\mathbf{P}}_o(t_s-\tau:t_s)|_n, \bar{\mathbf{P}}_o(t_e:t_e+\tau)|_n]),
\end{align}\end{linenomath}
where $t_s$ and $t_e$ are the starting and ending snippet indexes of the given TAP. $\bar{\mathbf{P}}_o(t_s:t_e)|_n$ indicates the fused prediction scores of the $n$-th action from the $t_s$-th snippet to the $t_e$-th snippet. $\tau=(t_e-t_s)/4$ denotes the inflation length and $\textrm{mean}(\cdot)$ is the mean function.

Instead of using Base module for TAL, our contribution is to leverage the output of the Subspace module (\ie, $\mathbf{P}_a$) to achieve better TAL results, as evaluated in Table~\ref{table:ablation_components}.
Without the attention mechanism in Subspace module, the TAP generation is achieved by thresholding the sum of all non-background classes ($\sum_{n=1}^{N}\bar{\mathbf{P}}_a|_n$) using $0.5$ ($C_2$ in Table~\ref{table:ablation_components}). For TAP evaluation, we use $\bar{\mathbf{P}}_a$ to replace $\bar{\mathbf{P}}_o$ in Eq. (\ref{eq:oic}) for better evaluation ($C_3$ in Table~\ref{table:ablation_components}) because $\bar{\mathbf{P}}_a$ can better avoid the distraction from the context.

\section{Experiments}
\begin{figure}[t]
  \begin{center}
  \includegraphics[width=0.46\textwidth]{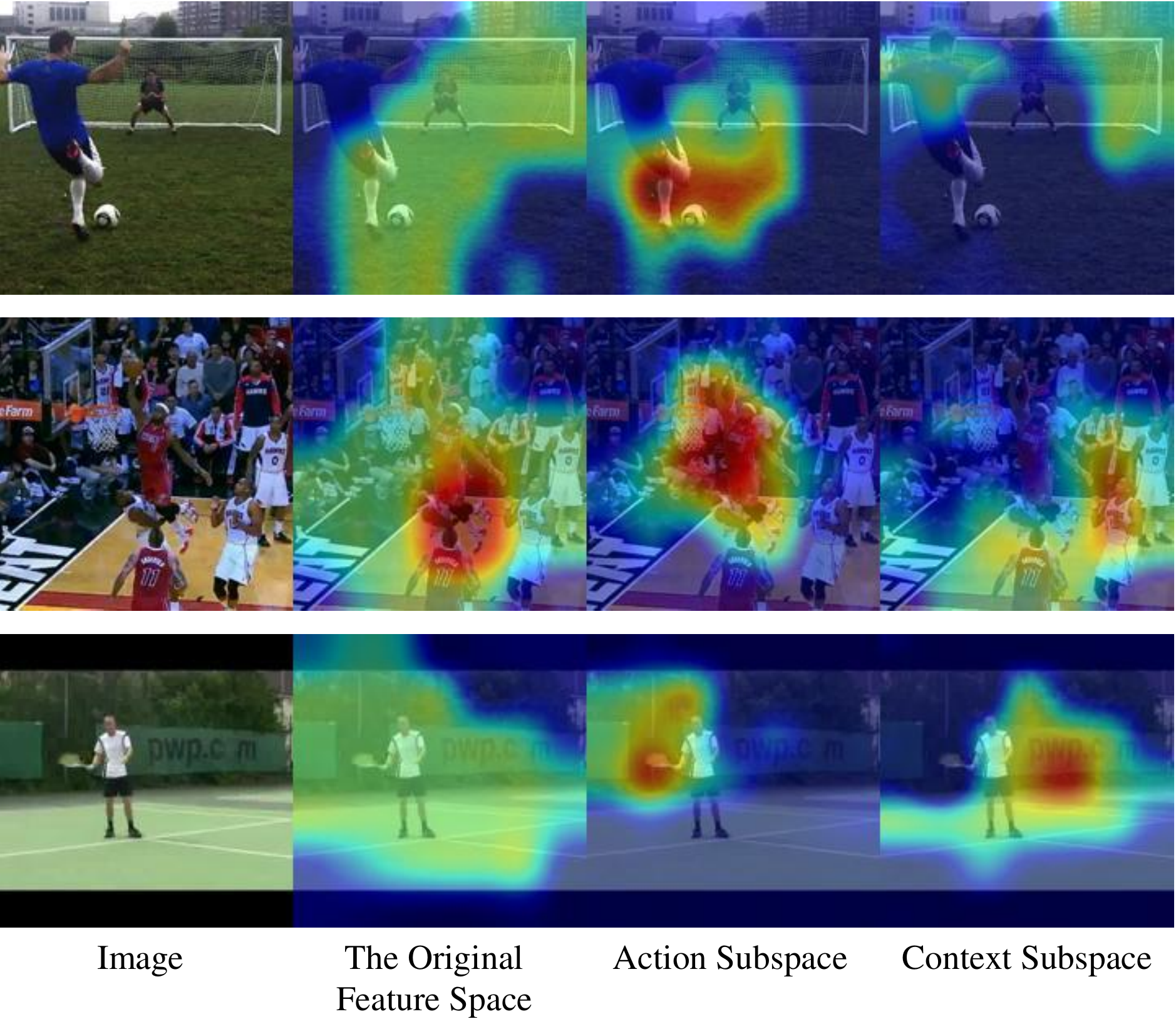}\\
  \caption{ The spacial visualization of the original feature space, action subspace and context subspace.
  The original feature space jointly represents both action and context visual elements.
  Compared with the context subspace, the action subspace focuses on the action areas (soccer penalty, basketball dunk and tennis swing) instead of context areas (soccer field, basketball court and tennis court).
  }\label{fig:gap}
  \end{center}
\end{figure}
\subsection{Experimental Setting}
\noindent\textbf{Evaluation Datasets.}
\textbf{THUMOS14.}~\cite{THUMOS14} We use the subset of the THUMOS14 dataset with temporal boundary annotations provided. It includes $20$ action categories. Following conventions, we train the proposed method using the validation set with 200 untrimmed videos and evaluate it on the test set with 213 untrimmed videos.
On average, each video contains $15.4$ action instances and $71.4\%$ frames are non-action background.
\textbf{ActivityNet.}~\cite{caba2015activitynet} The v1.2/v1.3 provides temporal boundary annotations for 100/200 activity classes with a total of $9,682$/$19,994$ videos. Since the temporal boundary annotations of the test set is not publicly available, we conduct experiments and report the performance on the validation set. The training set includes $4,819$/$10,024$ untrimmed videos and the validation set includes $2,383$/$4,926$ untrimmed videos. On average, each video contains $1.5$/$1.6$ action instances and $34.6\%$/$35.7\%$ non-action background, which indicates a relatively lower noise ratio compared with THUMOS14.

\noindent\textbf{Evaluation metric.}
Following conventions, we evaluate the TAL performance by mean average precision (mAP) values at different IoU thresholds. For the evaluation on THUMOS14, the IoU thresholds are $[0.1 : 0.1 : 0.9]$. For ActivityNet, the IoU thresholds are $[0.5 : 0.05 : 0.95]$.
\subsection{Ablation Study}
\begin{table}[t]\small
	\centering
		\caption{Four notations indicating different experiment settings used for ablation studies. $C_1$ is defined for comparison with Based module. $C_2$ and $C_3$ respectively represent the contribution of the Subspace module towards TAP generation and evaluation. $C_4$ is defined to solely evaluate the effect of T-ResM.}
		\label{table:parts}
\begin{tabular}{c|c}
\hline
{} &{Explanation} \\
\hline
$C_1$ & Thresholding $\bar{\mathbf{a}}$ with $0.5$ for TAP generation. \\
$C_2$ & Thresholding $\sum_{n=1}^{N}\bar{\mathbf{P}}_a|_n$ with $0.5$ for TAP generation.\\
$C_3$ & Using $\bar{\mathbf{P}}_a$ to replace $\bar{\mathbf{P}}_o$ in Eq. (\ref{eq:oic}) for TAP evaluation.\\
$C_4$ & Using T-ResM to enhance $\mathbf{F}$ following Eq. (\ref{eq:t_res}).\\
\hline
\end{tabular}
\end{table}

\begin{table}[t]\small
\caption{
Sensitivity test of hyper-parameters $\alpha$ (in Eq. (\ref{eq:gap})) and $\beta$ in Eq. (\ref{eq:beta0}-\ref{eq:beta1}) on THUMOS14 test set.
}
\begin{center}\small
\begin{tabular}{c|c|ccccc|c}
\hline
\multirow{2}{*}{$\alpha$}&\multirow{2}{*}{$\beta$}&\multicolumn{5}{c|}{mAP(\%)@IoU} & \multirow{2}{*}{AVG}\\
&& 0.3 & 0.4 & 0.5 & 0.6 & 0.7 &\\ \hline
\multicolumn{2}{c|}{0\#($\beta\!=\!0.4$)} & {38.4} & {30.5} & {21.5} & {14.5} & {7.3} & {22.4} \\
\multicolumn{2}{c|}{0\#($\beta\!=\!0.5$)} & {31.4} & {23.4} & {15.8} & {9.5} & {4.8} & {17.0} \\
\multicolumn{2}{c|}{0\#($\beta\!=\!0.6$)} & {27.2} & {20.3} & {13.6} & {7.6} & {4.0} & {14.5} \\ \hline\hline
\multirow{3}{*}{\shortstack{$0.1$}}
&$0.4$      & {50.6} & \textbf{41.7} & {29.7} & {20.2} & {10.4} & {30.5} \\
&$0.5$      & {49.7} & {41.3} & {30.1} & {19.6} & {10.6} & {30.3} \\
&$0.6$      & {49.9} & {41.0} & {29.9} & {19.7} & {10.1} & {30.1} \\
\hline
\multirow{3}{*}{\shortstack{$0.2$}}
&$0.4$       & \textbf{50.8} & \textbf{41.7} & {29.6} & {20.1} & {10.7} & \textbf{30.6} \\
&$0.5$       & {50.4} & \textbf{41.7} & \textbf{30.3} & {20.0} & {10.4} & {30.5} \\
&$0.6$       & {48.8} & {40.3} & {29.0} & {19.8} & {9.8} & {29.5} \\
\hline
\multirow{3}{*}{\shortstack{$0.3$}}
&$0.4$       & {50.2} & {41.4} & {29.9} & {20.0} & {10.6} & {30.4} \\
&$0.5$       & {48.8} & {40.8} & {30.1} & {20.0} & {10.1} & {30.0} \\
&$0.6$       & {49.2} & {39.5} & {28.3} & {18.6} & {9.3} & {29.0} \\
\hline
\multirow{3}{*}{\shortstack{$0.4$}}
&$0.4$       & {49.7} & {41.6} & {30.0} & \textbf{20.6} & {10.5} & {30.5} \\
&$0.5$       & {47.6} & {40.3} & {29.7} & {20.3} & \textbf{10.9} & {29.8} \\
&$0.6$       & {47.9} & {38.5} & {27.5} & {17.6} & {9.7} & {28.3} \\
\hline
\end{tabular}
\end{center}
\label{table:hp}
\end{table}
\subsubsection{Spatial Visualization of Action Subspace.}

To validate the learned action subspace focuses on the action visual elements, we visualize the predictions in it spatially.
To achieve the spatial visualization, we use the features before global average pooling and obtain $\bar{\mathbf{P}}_o$, $\bar{\mathbf{P}}_a$ and $\bar{\mathbf{P}}_c$ for every spatial location on the feature maps.
Facilitated by snippets with only context (\ie, $\mathbb{S}_c$) as negative training samples, $\bar{\mathbf{P}}_a$ can distinguish action from context.
However, due to actions seldom appear without their context spatially, our method is lack of snippets with \textbf{only} action (\eg, snippets with action soccer penalty but without a soccer field), which makes $\bar{\mathbf{P}}_c$ cannot directly distinguish context from action.
To eliminate action elements in $\bar{\mathbf{P}}_c$, we use $\textrm{max}(\bar{\mathbf{P}}_c-\bar{\mathbf{P}}_a,0)$ for every spatial location to illustrate the context subspace.
As shown in Figure~\ref{fig:gap}, the outputs of different feature spaces are visualized by heat maps imposed on the original images.

\subsubsection{Effect of Each Component on TAL Task.}
To evaluate the contribution of the proposed Subspace module, we first define notations indicating different experiment settings as listed in Table~\ref{table:parts}.
The effect of the proposed Subspace module on the TAL task is reflected in two aspects, \ie, using $\bar{\mathbf{P}}_a$ to improve both TAP generation and TAP evaluation steps ($C_2$ and $C_3$ in Table~\ref{table:parts}). Comparing 1\# (or 2\#) with 0\#, we can solely evaluate the contribution of $C_3$ (or $C_2$). Combining $C_2$ and $C_3$, the major improvement (48.8\% in UNT and 78.5\% in I3D) is achieved against the Base module. Facilitated by the T-ResM with the proposed unsupervised training task, the TAL performance can be further improved, as validated by the comparison between 3\# and 4\#. Compared with using I3D features, T-ResM plays a more important role when using UNT features, due to UNT features contain less temporal information compared with I3D features.
Our proposed components (\ie, $C_2$, $C_3$ and $C_4$) contribute to most of the performance gain (89.5\% in UNT and 90.8\% in I3D). Finally, with all the components, 5\# achieves the best performance with both feature backbones.

\subsubsection{Hyper-parameter Sensitivity.}

To explore the impact of the hyper-parameters, we adopt different settings of $\alpha$ and $\beta$, as summarized in Table~\ref{table:hp}.
The Base module only depends on $\beta$. Compared with results from Base module, our method, strengthened by the proposed Subspace module, shows robustness towards all hyper-parameters.

\begin{table}[t]\small
	\centering
		\caption{TAL performance comparison on ActivityNet v1.2 and v1.3 validation set, in terms of average mAP at IoU thresholds $[0.5 : 0.05 : 0.95].$ All results are obtained using I3D features.}
		\label{table:res_anet}
\resizebox{.47\textwidth}{!}{
\begin{tabular}{c|c|ccc|c}
\hline
\multirow{2}{*}{Method} & \multirow{2}{*}{\footnotesize{v1.2 / v1.3}} & \multicolumn{3}{c|}{mAP(\%)@IoU} & \multirow{2}{*}{Avg} \\
  & &  0.5    & 0.75& 0.95  &               \\\hline

 TSM ~(\citeyear{yu2019temporal})                 &v1.2 & {28.3} & {17.0} & {3.5} & {17.1} \\
{\lzynet}(\citeyear{lzyiccv})                  &v1.2 & {37.1} & {20.3} & {5.0} & {21.6} \\
CMCS (\citeyear{Liu_2019_CVPR})            &v1.2 & {36.8} &  22.0  & {5.6} & {22.4} \\
RPNet~(\citeyear{huang2020relational}) &v1.2 & {37.6} &  23.9  & {5.4} & {23.3} \\
\textbf{Ours}                      &v1.2& \textbf{39.2} & \textbf{25.6}& \textbf{6.8} & \textbf{25.5} \\\hline \hline
 TSM    ~(\citeyear{yu2019temporal})               &v1.3 & 30.3 & 19.0& 4.5 & -          \\
 CMCS  (\citeyear{Liu_2019_CVPR})           &v1.3 & {34.0} & 20.9 & \textbf{5.7} & {21.2} \\
 BM  ~(\citeyear{nguyen2019weakly})      &v1.3 & \textbf{36.4} & 19.2& 2.9 & -          \\
 BaSNet    ~(\citeyear{lee2019background})         &v1.3 & 34.5& 22.5& 4.9 & 22.2          \\
 \textbf{Ours}                      &v1.3 & 35.1 & \textbf{23.7} & {5.6} & \textbf{23.2}          \\
\hline
\end{tabular}
}
\end{table}

\begin{table*}[t]\small
\caption{
Ablation studies of our method on the THUMOS14 dataset by using mAP under different IoU thresholds and the average mAP under the IoU thresholds from 0.3 to 0.7. Notations indicating experiment settings are defined in Table~\ref{table:parts}. UNT and I3D represent UntrimmedNet and I3D feature backbones, respectively. The gain of average mAP against the Base module (0\#) of every variants are listed. Our proposed components (\ie, $C_2$, $C_3$ and $C_4$) contribute most of the performance gain.
}
\begin{center}
\resizebox{.94\textwidth}{!}{
\begin{tabular}{c|c|cccc|ccccccccc|cc}
\hline
{Ablated}&\multirow{2}{*}{Feature}&\multirow{2}{*}{$\!C_1\!$}&\multirow{2}{*}{$\!C_2\!$}&\multirow{2}{*}{$\!C_3\!$} &\multirow{2}{*}{$\!C_4\!$} &\multicolumn{9}{c|}{mAP(\%)@IoU} &{AVG} &\multirow{2}{*}{Gain}\\
{Variants}&&&&&& 0.1 & 0.2 & 0.3 & 0.4 & 0.5 & 0.6 & 0.7 &0.8 &0.9 &(0.3:0.7) &{}\\ \hline
0\#&UNT&\checkmark&&&& {50.0} & {43.6} & {35.7} & {29.0} & {21.7} & {14.2} & {7.4} & {2.5} & {0.3} & {21.6} & {-} \\
1\#&UNT&\checkmark&&\checkmark&& {51.3} & {45.4} & {37.3} & {30.4} & {22.5} & {15.0} & {7.8} & {2.7} & {0.3} & {22.6} & ${1.0|_{21.1\%}}$ \\
2\#&UNT&&\checkmark&&& {50.9} & {45.1} & {38.3} & {30.0} & {22.9} & {15.4} & {7.7} & {2.4} & {0.3} & {22.8} & $1.2|_{26.2\%}$ \\
3\#&UNT&&\checkmark&\checkmark&& {52.8} & {47.6} & {40.6} & {31.2} & {23.5} & {15.9} & {8.2} & {2.3} & {0.2} & {23.9} & $2.3|_{48.8\%}$ \\
4\#&UNT&&\checkmark&\checkmark&\checkmark& \textbf{54.8} & {48.9} & {41.8} & {33.7} & {26.1} & {18.0} & {9.4} & {3.4} & \textbf{0.5} & {25.8} & $4.2|_{89.5\%}$ \\
5\#&UNT&\checkmark&\checkmark&\checkmark&\checkmark& {54.7} & \textbf{49.1} & \textbf{42.1} & \textbf{34.2} & \textbf{26.7} & \textbf{18.5} & \textbf{9.7} & \textbf{3.5} & \textbf{0.5} & \textbf{26.3} & $4.6|_{100\%}$ \\ \hline \hline
0\#&I3D&\checkmark&&&& {48.9} & {44.7} & {38.4} & {30.5} & {21.5} & {14.5} & {7.3} & {2.6} & {0.3} & {22.4} & - \\
1\#&I3D&\checkmark&&\checkmark&& {58.9} & {54.8} & {47.4} & {38.3} & {26.9} & {17.7} & {8.9} & {3.2} & {0.3} & {27.9} & $5.5|_{66.6\%}$ \\
2\#&I3D&&\checkmark&&& {51.6} & {47.9} & {41.4} & {33.1} & {23.5} & {15.3} & {8.8} & {3.4} & \textbf{0.5} & {24.4} & $2.0|_{24.5\%}$ \\
3\#&I3D&&\checkmark&\checkmark&& {60.2} & {56.4} & {49.2} & {40.1} & {27.6} & {17.7} & {9.6} & {3.2} & \textbf{0.5} & {28.8} & $6.4|_{78.5\%}$ \\
4\#&I3D&&\checkmark&\checkmark&\checkmark& \textbf{62.2} & \textbf{58.2} & {50.7} & {40.6} & {28.3} & {19.2} & {10.4} & {4.0} & \textbf{0.5} & {29.8} & $7.4|_{90.8\%}$ \\
5\#&I3D&\checkmark&\checkmark&\checkmark&\checkmark& {61.7} & {58.0} & \textbf{50.8} & \textbf{41.7} & \textbf{29.6} & \textbf{20.1} & \textbf{10.7} & \textbf{4.3} & \textbf{0.5} & \textbf{30.6} & $8.2|_{100\%}$ \\ \hline
\end{tabular}
}
\end{center}
\label{table:ablation_components}
\end{table*}

\begin{table}[t]\small
\caption{Results on the THUMOS14 test set. We report mAP values at IoU thresholds 0.3:0.1:0.7. Recent works in both full and weak supervision settings are reported. Our method outperforms the state-of-the-art methods on both backbone settings.
}
\label{table:res_th}
\begin{center}
\resizebox{.48\textwidth}{!}{
\begin{tabular}{c|c|c|p{0.5cm}<{\centering}p{0.5cm}<{\centering}p{0.5cm}<{\centering}p{0.5cm}<{\centering}p{0.5cm}<{\centering}|c}
\hline
\multirow{2}{*}{} & \multirow{2}{*}{Method} & \multirow{2}{*}{\footnotesize{Feature}} & \multicolumn{5}{c|}{mAP@IoU} &  \multirow{2}{*}{AVG}\\
 & & & 0.3 & 0.4 & 0.5 & 0.6 & 0.7 &\\
\hline\hline
\multirow{2}{*}{\rotatebox{90}{\shortstack{Full}}}
&SSN~(\citeyear{zhao2017temporal})  & UNT & 51.9 & 41.0 & 29.8 & 19.6 & 10.7 & 30.6\\
&MGG~(\citeyear{MGG})   & I3D & 53.9 & 46.8 & 37.4 & 29.5 & 21.3 & 37.8\\
\hline \hline
\multirow{4}{*}{\rotatebox{90}{\shortstack{Weak}}}
&AutoLoc~(\citeyear{shou2018autoloc}) & UNT & 35.8 & 29.0 & 21.2 & 13.4 & 5.8     & 21.0\\
&CleanNet~(\citeyear{lzyiccv})        & UNT & 37.0 & 30.9 & 23.9 & 13.9 & 7.1     & 22.6\\
&RPNet~(\citeyear{huang2020relational})& UNT& 37.8 & 29.4 & 21.2 & 13.9 & 6.8  & 21.8\\
&\textbf{Ours}                & UNT & \textbf{42.1} & \textbf{34.2} & \textbf{26.7} & \textbf{18.5} & \textbf{9.7} & \textbf{26.3}\\ \hline\hline
\multirow{6}{*}{\rotatebox{90}{\shortstack{Weak}}}
&CMCS(\citeyear{Liu_2019_CVPR})& I3D & 41.2 & 32.1 & 23.1 & 15.0 & 7.0 & 23.7\\
&BM~(\citeyear{nguyen2019weakly}) & I3D &46.6 &37.5 &26.8 &17.6 &9.0  &27.5\\
&ASSG~(\citeyear{zhang2019adversarial})     & I3D &50.4 &38.7 &25.4 &15.0 &6.6  &27.2\\
&BaSNet~(\citeyear{lee2019background})       & I3D &44.6 &36.0 &27.0 &18.6 &10.4 &27.3\\
&RPNet~(\citeyear{huang2020relational})& I3D &48.2 &37.2 &27.9 &16.7 &8.1 & 27.6\\
&\textbf{Ours}               & I3D & \textbf{50.8} & \textbf{41.7} & \textbf{29.6} & \textbf{20.1} & \textbf{10.7} &\textbf{30.6}\\
\hline
\end{tabular}
}
\end{center}
\end{table}

\subsection{Comparisons with State-of-the-Art Methods}
As presented in Table~\ref{table:res_th}, our method significantly outperforms existing WS-TAL methods with both feature backbones on THUMOS14.
As presented in Table~\ref{table:res_anet}, our method also achieves the best average mAP compared with previous WS-TAL methods on ActivityNet v1.2 and v1.3.
However, the performance improvement is not as significant as that on THUMOS14, possibly due to the lower non-action frame rate in ActivityNet v1.2 and v1.3.
With less distraction from non-action context, the improvement brought by our method will be less significant.

In summary, the effectiveness of our method is validated by extensive experiments on different feature backbones and benchmarks.
Via spatial visualization, we validate that the original feature space indeed jointly represents both action and context visual elements. While in the learned action subspace, action visual elements are separated from context visual elements.
Through detailed ablation studies, the contribution of the proposed Subspace module is evaluated, and the robustness towards hyper-parameters is validated. Finally, we have compared our method with state-of-the-art WS-TAL methods on three standard benchmarks, and significant improvement is observed.

\section{Conclusion}

We propose to address the action-context confusion challenge for WS-TAL, by learning explicit action and context subspaces.
Leveraging the predictions from spatial and temporal streams for snippets grouping and introducing an auxiliary unsupervised training task, the two subspaces are learned and the distraction from the context is better avoided during temporal localization. Our method significantly outperforms existing \wtal~methods on three standard datasets. Visualization results and detailed ablation studies further validate the contribution of the proposed method.

\section{Acknowledgment}
This work was supported partly by National Key R\&D Program of China Grant 2018AAA0101400, NSFC Grants 61629301, 61773312, and 61976171, China Postdoctoral Science Foundation Grant 2019M653642, Young Elite Scientists Sponsorship Program by CAST Grant 2018QNRC001, and Natural Science Foundation of Shaanxi Grant 2020JQ-069.

\bibliography{aaai2021}
\end{document}